\documentclass[journal]{IEEEtran}

\usepackage{cite}
\usepackage{amsmath,amssymb,amsfonts}
\usepackage{algorithmic}
\usepackage{graphicx}
\usepackage{textcomp}
\usepackage{xcolor}
\usepackage{booktabs}
\usepackage{multirow}
\usepackage{array}
\usepackage{url}
\usepackage{hyperref}
\usepackage{balance}
\usepackage{subcaption}

\hypersetup{
    colorlinks=true,
    linkcolor=blue,
    citecolor=blue,
    urlcolor=blue
}

\begin{document}

\title{LUT-Compiled Kolmogorov-Arnold Networks for\\Lightweight DoS Detection on IoT Edge Devices}

\author{Oleksandr~Kuznetsov$^{\dagger}$,~\IEEEmembership{Member,~IEEE}

\thanks{O. Kuznetsov is with the Department of Theoretical and Applied Sciences, eCampus University, Via Isimbardi 10, Novedrate (CO), 22060, Italy, and also with the Department of Intelligent Software Systems and Technologies, School of Computer Science and Artificial Intelligence, V.N. Karazin Kharkiv National University, 4 Svobody Sq., 61022 Kharkiv, Ukraine. E-mail: oleksandr.kuznetsov@uniecampus.it, kuznetsov@karazin.ua}
\thanks{$^{\dagger}$Corresponding author: oleksandr.kuznetsov@uniecampus.it}
} 

\maketitle

\begin{abstract}
Denial-of-Service (DoS) attacks pose a critical threat to Internet of Things (IoT) ecosystems, yet deploying effective intrusion detection on resource-constrained edge devices remains challenging. Kolmogorov-Arnold Networks (KANs) offer a compact alternative to Multi-Layer Perceptrons (MLPs) by placing learnable univariate spline functions on edges rather than fixed activations on nodes, achieving competitive accuracy with fewer parameters. However, runtime B-spline evaluation introduces significant computational overhead unsuitable for latency-critical IoT applications. We propose a lookup table (LUT) compilation pipeline that replaces expensive spline computations with precomputed quantized tables and linear interpolation, dramatically reducing inference latency while preserving detection quality. Our lightweight KAN model (50K parameters, 0.19~MB) achieves 99.0\% accuracy on the CICIDS2017 DoS dataset. After LUT compilation with resolution $L=8$, the model maintains 98.96\% accuracy (F1 degradation $<0.0004$) while achieving $\mathbf{68\times}$ speedup at batch size 256 and over $\mathbf{5000\times}$ speedup at batch size 1, with only $2\times$ memory overhead. We provide comprehensive evaluation across LUT resolutions, quantization schemes, and out-of-bounds policies, establishing clear Pareto frontiers for accuracy-latency-memory trade-offs. Our results demonstrate that LUT-compiled KANs enable real-time DoS detection on CPU-only IoT gateways with deterministic inference latency and minimal resource footprint.
\end{abstract}


\section{Introduction}
\label{sec:introduction}

The proliferation of Internet of Things (IoT) devices has created vast attack surfaces vulnerable to Denial-of-Service (DoS) attacks~\cite{kolias2017ddos}. Unlike traditional computing environments, IoT ecosystems comprise heterogeneous devices with stringent resource constraints---limited memory, CPU-only processing, and strict power budgets---making conventional deep learning approaches impractical for real-time threat detection~\cite{diro2018distributed}.

Machine learning-based intrusion detection systems (IDS) have demonstrated remarkable accuracy in identifying network anomalies~\cite{rajathi2025hybrid}. However, state-of-the-art models often require substantial computational resources: SecEdge~\cite{awan2025secedge} achieves 98.7\% accuracy on CICIDS2017, but demands 1.1--1.7~GB memory, while ALNS-CNN~\cite{cherfi2024alns} requires high CPU utilization for its accelerated inference. This creates a fundamental tension between detection capability and deployability on edge devices.

Kolmogorov-Arnold Networks (KANs)~\cite{liu2024kan} offer a promising alternative by leveraging the Kolmogorov-Arnold representation theorem~\cite{kolmogorov1957representation}. Unlike Multi-Layer Perceptrons (MLPs) that apply fixed activation functions at nodes, KANs place learnable univariate functions---typically implemented as B-splines---on network edges. This architectural difference enables KANs to achieve competitive accuracy with significantly fewer parameters and enhanced interpretability~\cite{sulaiman2024kan}. Recent applications in network security~\cite{wang2025kinn,wu2025d2kco} have demonstrated KANs' effectiveness for intrusion detection tasks.

Despite their parameter efficiency, KANs suffer from a critical deployment bottleneck: runtime B-spline evaluation requires iterative knot interval search, recursive basis function computation, and coefficient aggregation for each input dimension---operations that dominate inference time on CPU architectures common in IoT gateways. This computational overhead undermines the practical advantages of KANs' compact representation.

\textbf{Our Contribution.} We address this deployment gap by proposing a lookup table (LUT) compilation pipeline that transforms trained KAN models for efficient edge inference. Our approach:
\begin{enumerate}
    \item \textbf{Eliminates runtime spline evaluation} by precomputing discretized spline values into quantized lookup tables
    \item \textbf{Replaces iterative algorithms} with simple table indexing and linear interpolation
    \item \textbf{Achieves 68$\times$ speedup} at batch size 256 and over 5000$\times$ at batch size 1 with minimal accuracy loss
    \item \textbf{Provides comprehensive evaluation} across LUT resolutions $L \in \{2,4,8,16,32,64,128,256\}$, quantization schemes, and boundary handling policies
\end{enumerate}

\begin{figure}[t]
    \centering
    \includegraphics[width=\columnwidth]{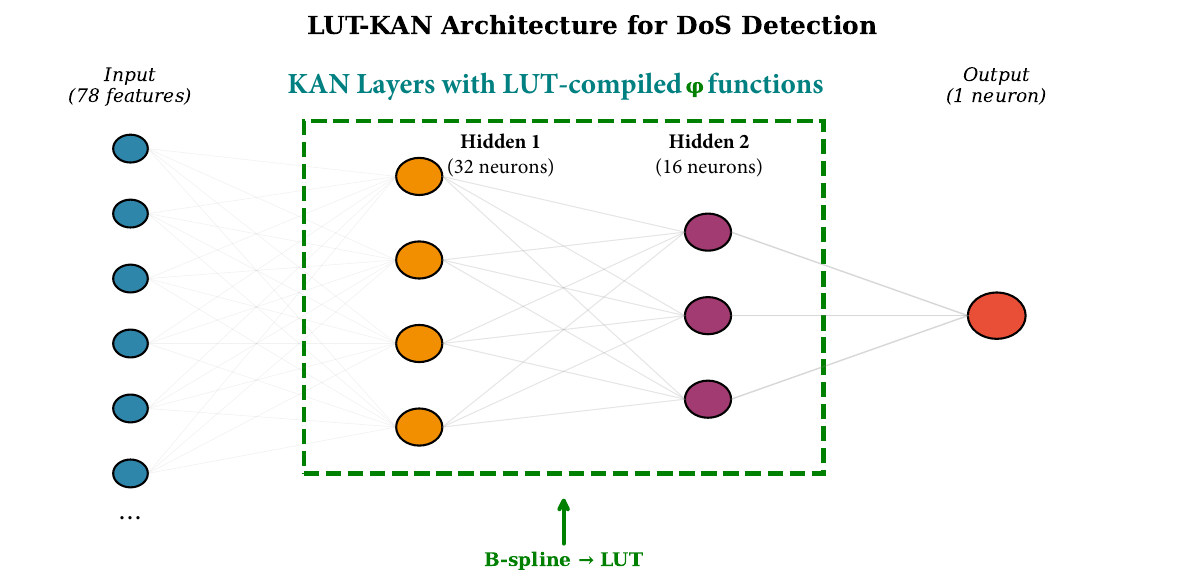}
    \caption{LUT-KAN architecture for DoS detection. The network processes 78 flow-level features through two KAN layers with 32 and 16 neurons respectively. Each learnable $\phi$ function is compiled from B-splines to precomputed lookup tables for efficient edge inference.}
    \label{fig:architecture}
\end{figure}

Fig.~\ref{fig:architecture} illustrates the LUT-KAN architecture. Our lightweight model (78 input features $\rightarrow$ [32, 16] hidden neurons $\rightarrow$ 1 output) achieves 99.0\% accuracy on the CICIDS2017 DoS dataset while occupying only 0.19~MB---suitable for deployment on resource-constrained IoT gateways.

\section{Related Work}
\label{sec:related}

\subsection{IoT Intrusion Detection Under Edge Constraints}

Deploying IDS on IoT edge devices requires balancing detection accuracy against strict resource budgets. Recent works have explored various approaches to this challenge. SecEdge~\cite{awan2025secedge} proposed a hybrid deep learning architecture achieving 98.7\% accuracy on CICIDS2017, but requires 1.1--1.7~GB memory. ALNS-CNN~\cite{cherfi2024alns} introduced accelerated learning with 99.85\% accuracy, though CPU utilization remains high. NIDS-DA~\cite{kumar2025nidsda} achieved 99.97\% accuracy with only 5K parameters through data augmentation, demonstrating that compact models can maintain high detection rates.

Edge deployment strategies include hybrid CNN-LSTM architectures~\cite{nazir2024cnnlstm}, ensemble methods~\cite{urmi2024stacked}, and metaheuristic optimization~\cite{bikila2025edagwo}. However, these approaches typically target MLP or CNN architectures without addressing the unique computational patterns of KAN-based models.

\subsection{Kolmogorov-Arnold Networks for Security}

KANs~\cite{liu2024kan} have recently emerged as an alternative to MLPs for various tasks. The theoretical foundation---the Kolmogorov-Arnold representation theorem---guarantees that any multivariate continuous function can be represented as a superposition of univariate functions. Practical implementations use B-splines to approximate these univariate functions, enabling gradient-based learning.

In security applications, DeepOKAN~\cite{abueidda2025deepokan} demonstrated KANs' potential for operator learning in mechanics problems, establishing architectural patterns applicable to network traffic analysis. KINN~\cite{wang2025kinn} applied KANs to physics-informed learning with competitive results. Recent work explored KAN applications in load forecasting~\cite{danish2025karn}, autonomous driving~\cite{huang2026kanidrl}, and computation offloading in Industrial IoT~\cite{wu2025d2kco}. However, none of these works address the inference efficiency challenge critical for IoT deployment.

\subsection{Quantization and LUT-Based Inference}

Post-training quantization~\cite{jacob2018quantization} reduces model size and accelerates inference by representing weights and activations with lower-precision integers. For specialized hardware, LUT-based approaches have shown remarkable efficiency. PolyLUT~\cite{andronic2023polylut} demonstrated FPGA implementations using polynomial approximations for ultra-low latency inference. NeuraLUT~\cite{andronic2024neuralut} explored hiding neural network density in Boolean synthesizable functions, achieving significant latency improvements through learned lookup tables.

Our work differs by targeting CPU-only IoT devices and specifically addressing KAN's B-spline evaluation bottleneck through systematic LUT compilation with comprehensive trade-off analysis.

\section{Background and Motivation}
\label{sec:background}

\subsection{IoT Edge Inference Constraints}

Intrusion Detection Systems for IoT networks operate under tighter constraints than conventional enterprise IDS. At the edge, inference must often run on CPU-only platforms with limited memory, energy budget, and strict latency constraints. In many realistic deployments, predictions are required in an online mode (near-real-time stream processing), where the effective batch size is small (often batch=1). In this regime, models that are accurate but require heavy runtime stacks, dynamic dispatch, or expensive non-linear primitives become impractical.

DoS attacks generate high-volume and fast-changing traffic patterns. The IDS must therefore provide both high detection quality and fast inference to avoid becoming a bottleneck or missing time-critical events. Key requirements include:
\begin{itemize}
    \item \textbf{Latency}: Sub-millisecond inference for individual flows
    \item \textbf{Memory}: Model size $<10$~MB for deployment alongside other services
    \item \textbf{Determinism}: Predictable inference time without garbage collection pauses
    \item \textbf{CPU-only}: GPU/TPU acceleration unavailable on typical IoT hardware
\end{itemize}

\subsection{KAN Architecture and B-Spline Formulation}

A KAN layer transforms input $\mathbf{x} \in \mathbb{R}^{n_{\text{in}}}$ to output $\mathbf{h} \in \mathbb{R}^{n_{\text{out}}}$ through:
\begin{equation}
    h_j = \sum_{i=1}^{n_{\text{in}}} \phi_{ij}(x_i), \quad j = 1, \ldots, n_{\text{out}}
    \label{eq:kan_layer}
\end{equation}
where each $\phi_{ij}$ is a learnable univariate function. In PyKAN-style implementations, $\phi_{ij}$ is modeled as a sum of a base branch and a spline branch:
\begin{equation}
    \phi_{ij}(x) = \alpha_{ij} \cdot b(x) + \beta_{ij} \cdot s_{ij}(x)
    \label{eq:phi_base_spline}
\end{equation}
where $b(\cdot)$ is a fixed nonlinear function (e.g., SiLU), $s_{ij}(\cdot)$ is a spline, and $\alpha_{ij}, \beta_{ij}$ are learned scaling factors.

The spline branch is represented via B-splines on a knot sequence. Let $\{t_u\}_{u=1}^{M}$ be an augmented knot vector and let $\{c_m\}_{m=1}^{P}$ be spline coefficients. The degree-$k$ B-spline basis functions $B_{m,k}(x)$ are defined recursively:
\begin{align}
    B_{m,0}(x) &= \mathbb{I}[t_m \le x < t_{m+1}] \label{eq:bspline_deg0} \\
    B_{m,k}(x) &= \frac{x - t_m}{t_{m+k} - t_m} B_{m,k-1}(x) + \frac{t_{m+k+1} - x}{t_{m+k+1} - t_{m+1}} B_{m+1,k-1}(x)
    \label{eq:bspline_recursive}
\end{align}

The spline value is computed as:
\begin{equation}
    s_{ij}(x) = \sum_{m=1}^{P} c^{(ij)}_{m} \cdot B_{m,k}(x)
    \label{eq:spline_sum}
\end{equation}

The full model is a composition of $L_{\text{layers}}$ such layers, followed by an output activation to produce a probability:
\begin{equation}
    \hat{p}(\mathbf{x}) = \sigma(f_{\theta}(\mathbf{x})), \qquad \hat{y} = \mathbb{I}[\hat{p}(\mathbf{x}) \ge \tau]
    \label{eq:prediction}
\end{equation}
where $\sigma(\cdot)$ is the logistic function, $\tau$ is the decision threshold (we use $\tau=0.5$), and $\theta$ denotes all model parameters.

\subsection{Computational Bottleneck Analysis}

For each input $x_i$, evaluating $s_{ij}(x_i)$ requires: (1) locating the knot interval containing $x_i$, (2) recursively computing $k+1$ non-zero basis functions, and (3) aggregating with coefficients. On CPU, this sequential computation dominates inference time---particularly at batch size 1 where vectorization benefits vanish.

This motivates a compilation view of KAN inference: if the learned univariate functions are fixed at deployment time, we can precompute their values on a grid and replace runtime spline evaluation with fast table lookup and interpolation.

\subsection{Problem Statement}

Let the trained KAN model define, for each layer $\ell$ and each edge $(i \rightarrow j)$, a univariate function $\phi^{(\ell)}_{ij}(x)$. Standard inference evaluates $\phi^{(\ell)}_{ij}$ via B-splines at runtime. We address the following deployment problem:

\emph{Given a trained KAN-based IDS model and a target CPU-only inference environment, how can we reduce inference latency while maintaining detection quality and controlling memory overhead?}

In addition, practical deployment requires robustness to out-of-domain inputs. Network telemetry may drift because of changes in traffic distributions, feature scaling issues, or sensor noise. Therefore, the LUT representation must define explicit behavior for out-of-bounds (OOB) inputs.

\section{Models and Methods}
\label{sec:methods}

\subsection{Dataset and Task Definition}

We use the CICIDS2017 dataset~\cite{sharafaldin2018toward} and focus on a DoS detection scenario. Each record corresponds to a network flow described by a vector of tabular features. The supervised task is binary classification: \emph{benign} vs. \emph{DoS}. Let $\mathbf{x} \in \mathbb{R}^d$ denote the feature vector and $y \in \{0,1\}$ the ground-truth label.

Table~\ref{tab:dataset} summarizes the dataset composition after preprocessing.

\begin{table}[h]
\centering
\caption{CICIDS2017 Wednesday DoS Dataset Statistics}
\label{tab:dataset}
\begin{tabular}{lrr}
\toprule
\textbf{Class} & \textbf{Samples} & \textbf{Percentage} \\
\midrule
Benign & 440,031 & 95.2\% \\
DoS Hulk & 231,073 & -- \\
DoS GoldenEye & 10,293 & -- \\
DoS slowloris & 5,796 & -- \\
DoS Slowhttptest & 5,499 & -- \\
Heartbleed & 11 & -- \\
\midrule
\textbf{Total Attack} & \textbf{252,672} & 4.8\% \\
\bottomrule
\end{tabular}
\end{table}

\textbf{Preprocessing.} Following established protocols~\cite{rajathi2025hybrid}, we: (1) remove constant and duplicate features, (2) apply the $3\sigma$ rule for outlier removal, (3) impute missing values with median, (4) standardize to zero mean and unit variance, and (5) balance classes through stratified sampling (231,073 samples per class). The final feature set comprises 78 flow-level statistics.

\begin{figure}[t]
    \centering
    \includegraphics[width=0.95\columnwidth]{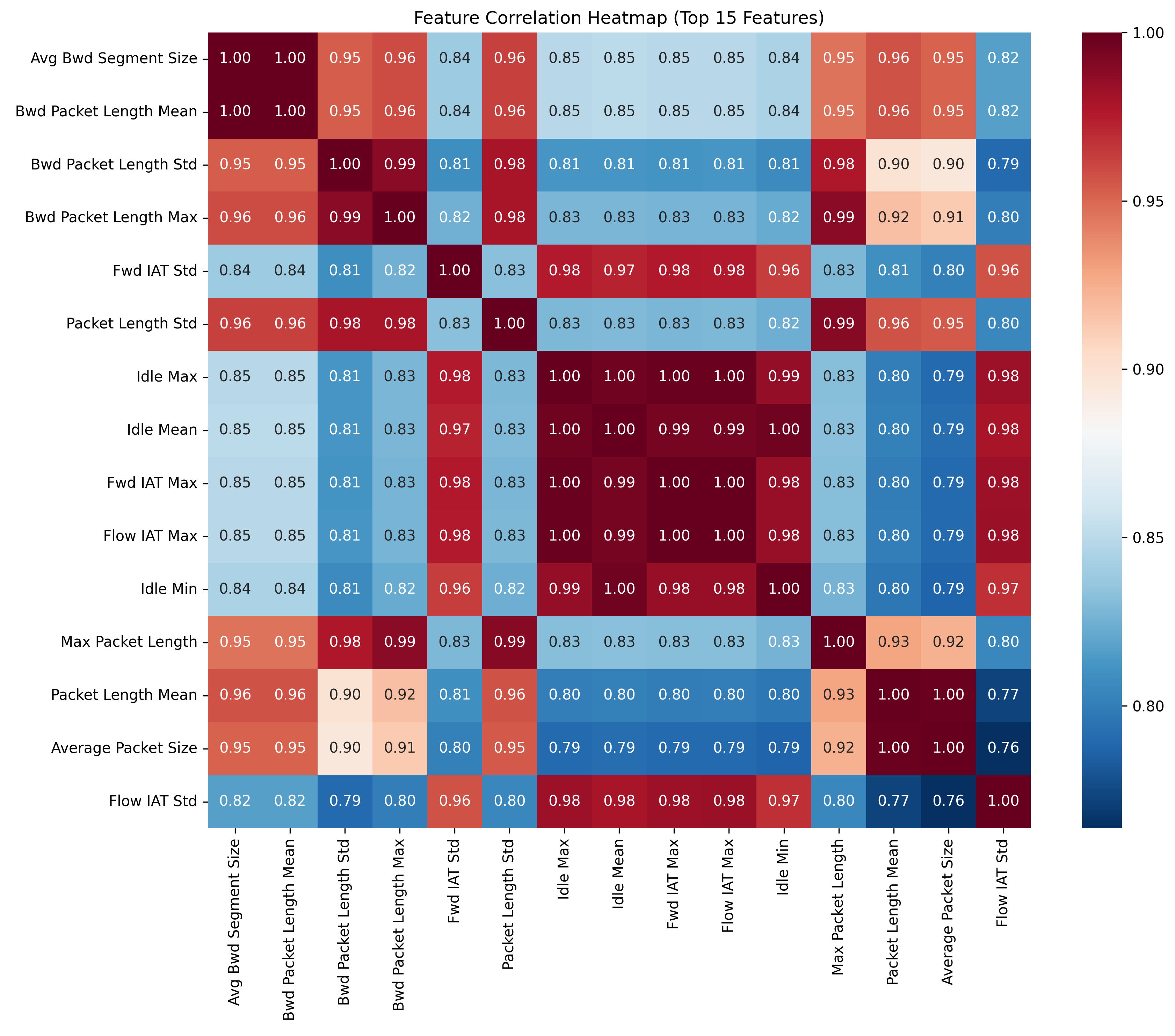}
    \caption{Feature correlation heatmap for top 15 features. Strong correlations exist within packet length statistics (upper-left cluster) and inter-arrival time features (center cluster).}
    \label{fig:correlation}
\end{figure}

Fig.~\ref{fig:correlation} shows the correlation heatmap for the top 15 most correlated features, revealing distinct clusters related to packet timing (IAT features) and size statistics.

\subsection{KAN Model Architecture and Training}

Our detection model employs a two-layer KAN architecture:
\begin{equation}
    \text{Input}(78) \rightarrow \text{KAN}_1(32) \rightarrow \text{KAN}_2(16) \rightarrow \text{Output}(1)
\end{equation}

Each KAN layer uses cubic B-splines ($k=3$) with $G=5$ grid intervals, yielding $G+k=8$ control points per spline. The model totals 50,092 parameters (42,336 trainable) occupying 0.19~MB in float32 representation.

\textbf{Training Procedure.} We train the model by minimizing the binary cross-entropy loss:
\begin{equation}
    \mathcal{L}(\theta) = -\frac{1}{N}\sum_{n=1}^{N} \left[ y_n \log \hat{p}(\mathbf{x}_n) + (1-y_n)\log (1-\hat{p}(\mathbf{x}_n)) \right]
    \label{eq:bce}
\end{equation}

\begin{figure}[t]
    \centering
    \includegraphics[width=\columnwidth]{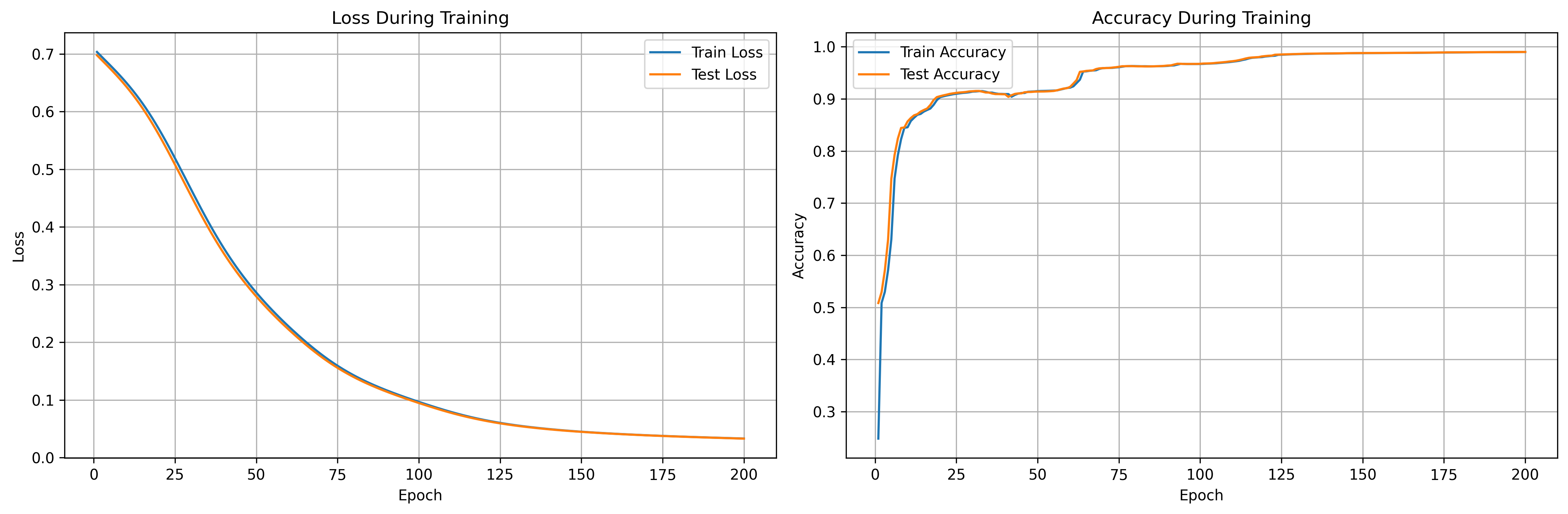}
    \caption{Training dynamics over 200 epochs. Left: Loss convergence showing stable optimization without overfitting. Right: Accuracy progression demonstrating rapid initial learning followed by gradual refinement.}
    \label{fig:training}
\end{figure}

Optimization is performed with AdamW optimizer (learning rate $10^{-3}$, weight decay $10^{-4}$) for 200 epochs with batch size 256. An 80/20 stratified train/test split ensures balanced evaluation. Fig.~\ref{fig:training} shows the training dynamics.

\subsection{LUT Compilation and Quantization}

After training, we compile the spline branch of each $\phi_{ij}$ into a lookup-table representation. For each spline segment, we sample $L$ points uniformly in the segment domain and precompute spline values. At inference time, for an input $x$, we: (i) identify the segment index, (ii) compute a normalized position within the segment, and (iii) perform linear interpolation between adjacent LUT values.

Let the segment domain be $[a,b]$ and let $\{v_q\}_{q=0}^{L-1}$ denote precomputed values at grid points $x_q = a + q\Delta$ with $\Delta = (b-a)/(L-1)$. For any $x \in [a,b]$, define $u = (x-a)/\Delta$, $q = \lfloor u \rfloor$, and $\lambda = u - q$. The interpolated value is:
\begin{equation}
    \tilde{s}_{ij}(x) = (1 - \lambda) \cdot v_q + \lambda \cdot v_{q+1}
    \label{eq:lut_interp}
\end{equation}

\textbf{Symmetric int8 Quantization.} We quantize LUT values per segment using:
\begin{equation}
    q = \mathrm{clip}\left(\left\lfloor \frac{v}{s} \right\rceil, -127, 127\right), \qquad \hat{v} = s \cdot q
    \label{eq:symmetric_quant}
\end{equation}
where $s$ is a segment-wise scale, and $\lfloor \cdot \rceil$ denotes rounding to nearest integer.

\textbf{Asymmetric uint8 Quantization.} For affine quantization:
\begin{equation}
    q = \mathrm{clip}\left(\left\lfloor \frac{v - v_{\min}}{s} \right\rceil, 0, 255\right), \qquad \hat{v} = v_{\min} + s \cdot q
    \label{eq:asymmetric_quant}
\end{equation}
where $v_{\min}$ and $s$ are stored per segment.

\subsection{Out-of-Bounds Policies and Boundary Modes}

The LUT representation must define behavior for inputs outside the trained spline domain. We consider two OOB policies:
\begin{itemize}
    \item \textbf{clip\_x}: clip $x$ into the valid domain and evaluate the boundary segment
    \item \textbf{zero\_spline}: return $0$ for the spline branch when $x$ is out-of-range
\end{itemize}

We also consider two boundary modes for segment indexing: \texttt{closed} (inclusive boundary) and \texttt{half\_open} (half-open intervals), which affect how exact knot boundary values are mapped to segments.

\subsection{Evaluation Metrics}

We report thresholded classification metrics at $\tau=0.5$: Accuracy, Precision, Recall, and F1. We also report threshold-free metrics: ROC-AUC and PR-AUC. Let TP, FP, TN, FN denote confusion matrix counts; then:
\begin{align}
    \mathrm{Precision} &= \frac{\mathrm{TP}}{\mathrm{TP}+\mathrm{FP}}, \qquad \mathrm{Recall} = \frac{\mathrm{TP}}{\mathrm{TP}+\mathrm{FN}} \label{eq:prec_rec} \\
    \mathrm{F1} &= \frac{2 \cdot \mathrm{Precision} \cdot \mathrm{Recall}}{\mathrm{Precision} + \mathrm{Recall}} \label{eq:f1}
\end{align}

Latency is reported as milliseconds per sample in two regimes: batch=1 and batch=256. All reported numbers are mean $\pm$ standard deviation across $n=5$ calibration seeds, with 95\% CIs computed as:
\begin{equation}
    \mathrm{CI}_{95\%} \approx \mu \pm t_{0.975,4} \cdot \frac{\sigma}{\sqrt{5}}, \quad t_{0.975,4} \approx 2.776
    \label{eq:ci}
\end{equation}

\section{Experimental Results}
\label{sec:results}

\subsection{Experimental Setup}

All experiments execute on a single AMD Ryzen 7 7840HS processor (8 cores, 3.8~GHz base) with 64~GB DDR5 RAM, running Windows 11 with Python 3.10. We report CPU-only performance to reflect IoT deployment conditions.

\textbf{Latency Measurement Protocol.} We measure \emph{infer-only} latency, which excludes data loading, preprocessing, and postprocessing. Each measurement consists of: (1) 10 warm-up iterations to ensure JIT compilation and cache stabilization, (2) 100 timed iterations, (3) reporting mean and standard deviation across 5 independent calibration seeds. All measurements use single-threaded execution (\texttt{NUMBA\_NUM\_THREADS=1}, \texttt{OMP\_NUM\_THREADS=1}) to simulate resource-constrained IoT environments. The baseline latency is measured using PyKAN's native B-spline evaluation with the same threading constraints.

\textbf{LUT Compilation Configuration.} Unless stated otherwise, LUT compilation uses \texttt{value\_repr=spline\_component}, \texttt{interp=linear}, and we sweep the LUT resolution $L \in \{2,4,8,16,32,64,128,256\}$. We evaluate both quantization schemes: symmetric int8 and asymmetric uint8. Main results use the robust setting (\texttt{half\_open} + \texttt{zero\_spline}).

\subsection{Baseline Model Performance}

\begin{figure}[t]
    \centering
    \includegraphics[width=\columnwidth]{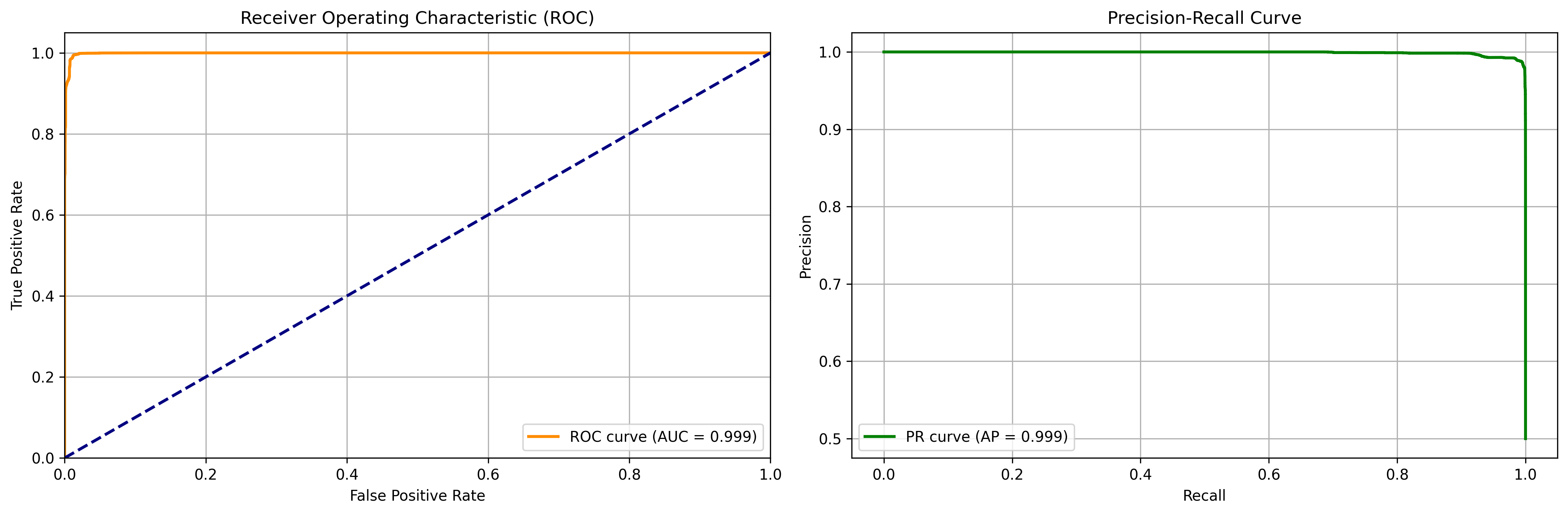}
    \caption{Baseline model evaluation curves. Left: ROC curve with AUC=0.999. Right: Precision-Recall curve with AP=0.999 demonstrating robust performance.}
    \label{fig:roc_pr}
\end{figure}

The float KAN model achieves strong detection performance, as shown in Table~\ref{tab:baseline}. Fig.~\ref{fig:roc_pr} shows the ROC and Precision-Recall curves demonstrating near-perfect classification capability.

\begin{table}[h]
\centering
\caption{Baseline KAN Model Performance}
\label{tab:baseline}
\begin{tabular}{lc}
\toprule
\textbf{Metric} & \textbf{Value} \\
\midrule
Accuracy & 99.0\% \\
Precision & 98.4\% \\
Recall & 99.6\% \\
F1 Score & 99.0\% \\
ROC-AUC & 0.999 \\
PR-AUC & 0.999 \\
\midrule
Parameters & 50,092 \\
Model Size & 0.19 MB \\
\bottomrule
\end{tabular}
\end{table}

\begin{figure}[t]
    \centering
    \includegraphics[width=\columnwidth]{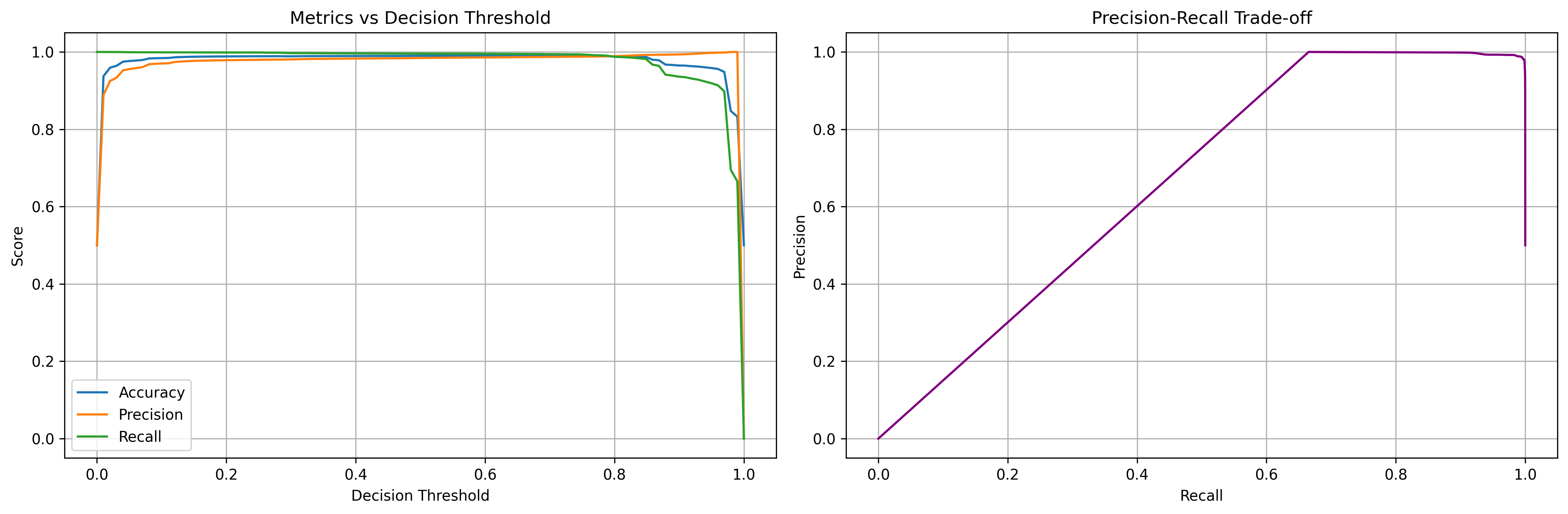}
    \caption{Model threshold analysis. (a) Performance metrics remain stable across decision thresholds. (b) Precision-recall trade-off curve.}
    \label{fig:threshold}
\end{figure}

Fig.~\ref{fig:threshold} illustrates model robustness across decision thresholds, with stable performance over a wide operating range ($\tau \in [0.1, 0.9]$).

\subsection{LUT Quality Preservation}

Table~\ref{tab:quality_full} presents comprehensive detection quality across LUT resolutions and quantization schemes. Even at $L=2$ (coarsest resolution), accuracy remains above 98.7\%. At $L=8$, the model achieves 98.96\% accuracy with F1 degradation of only 0.0004 from the float baseline.

\begin{table}[t]
\centering
\caption{Detection Quality on Test Split (mean across $n=5$ runs; std $<10^{-4}$ for all configurations). Baseline: Acc=0.9899, F1=0.9900.}
\label{tab:quality_full}
\scriptsize
\begin{tabular}{rlcccc}
\toprule
$L$ & Quant. & Accuracy & F1 & ROC-AUC & $\Delta$F1 \\
\midrule
2   & sym int8   & 0.9874 & 0.9874 & 0.9991 & -0.0026 \\
2   & asym uint8 & 0.9873 & 0.9874 & 0.9991 & -0.0026 \\
4   & sym int8   & 0.9886 & 0.9887 & 0.9991 & -0.0013 \\
4   & asym uint8 & 0.9886 & 0.9887 & 0.9991 & -0.0013 \\
\textbf{8}   & \textbf{sym int8}   & \textbf{0.9895} & \textbf{0.9896} & \textbf{0.9991} & \textbf{-0.0004} \\
8   & asym uint8 & 0.9895 & 0.9896 & 0.9991 & -0.0004 \\
16  & sym int8   & 0.9896 & 0.9897 & 0.9991 & -0.0003 \\
32  & sym int8   & 0.9897 & 0.9898 & 0.9991 & -0.0003 \\
64  & sym int8   & 0.9898 & 0.9898 & 0.9991 & -0.0002 \\
128 & sym int8   & 0.9899 & 0.9899 & 0.9991 & -0.0001 \\
256 & sym int8   & 0.9900 & 0.9900 & 0.9991 & $<$0.0001 \\
\bottomrule
\end{tabular}
\end{table}

\begin{figure}[t]
    \centering
    \includegraphics[width=0.85\columnwidth]{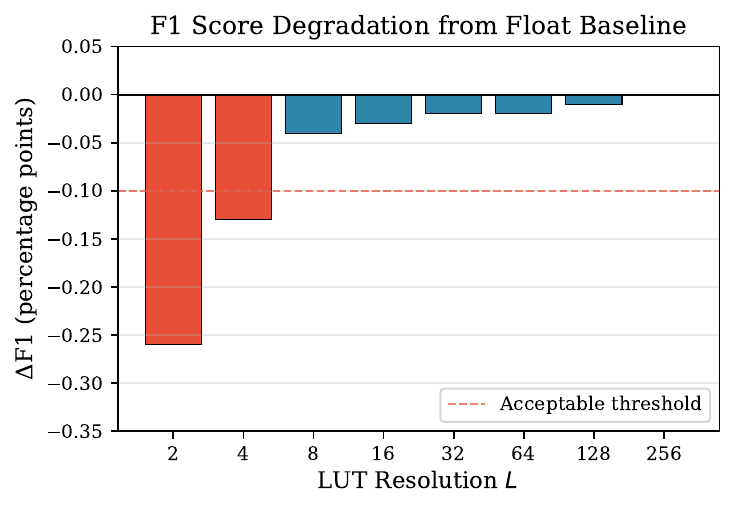}
    \caption{F1 score degradation from float baseline across LUT resolutions. All configurations with $L \geq 4$ remain within the acceptable threshold.}
    \label{fig:f1_degradation}
\end{figure}

Fig.~\ref{fig:f1_degradation} visualizes F1 degradation across resolutions.

\subsection{Inference Latency Analysis}

\textbf{Fair Baseline Comparison.} To ensure valid speedup claims, we compare backends with matched implementations. Table~\ref{tab:fair_baseline} presents the fair baseline comparison at $L=8$, where both B-spline evaluation and LUT inference use the same Numba JIT-compiled backend. This eliminates confounding factors from Python interpreter overhead and library differences.

\begin{table}[t]
\centering
\caption{Fair Baseline Comparison at $L=8$ (Numba backend for both)}
\label{tab:fair_baseline}
\scriptsize
\begin{tabular}{lcccc}
\toprule
\textbf{Method} & \textbf{Batch=1} & \textbf{Batch=256} & \textbf{Speedup (bs=1)} & \textbf{Speedup (bs=256)} \\
\midrule
Numba B-spline & 158.9 ms & 0.878 ms & 1.0$\times$ & 1.0$\times$ \\
Numba LUT (int8) & 0.025 ms & 0.013 ms & 6333$\times$ & 68$\times$ \\
\bottomrule
\end{tabular}
\end{table}

The dramatic speedup difference between batch sizes reflects the nature of the computational bottleneck: at batch=1, PyKAN's B-spline evaluation suffers from Python dispatch overhead and lack of vectorization, while LUT lookup remains efficient. At batch=256, vectorization partially amortizes the B-spline overhead, reducing the relative speedup.

Table~\ref{tab:latency_batch256} and Table~\ref{tab:latency_batch1} present inference latency with 95\% confidence intervals for batch sizes 256 and 1, respectively.

\begin{table}[t]
\centering
\caption{Infer-only Latency, Batch=256 (Numba LUT). Baseline: 0.878$\pm$0.097 ms/sample.}
\label{tab:latency_batch256}
\scriptsize
\begin{tabular}{rlcccc}
\toprule
$L$ & Quant. & ms/sample ($\mu\pm\sigma$) & CI$_{95\%}$ & Speedup \\
\midrule
2   & asym uint8 & 0.0129$\pm$0.0032 & $\pm$0.0040 & 68.0$\times$ \\
2   & sym int8   & 0.0142$\pm$0.0043 & $\pm$0.0053 & 61.3$\times$ \\
4   & asym uint8 & 0.0130$\pm$0.0006 & $\pm$0.0008 & 64.7$\times$ \\
4   & sym int8   & 0.0135$\pm$0.0007 & $\pm$0.0009 & 61.7$\times$ \\
\textbf{8}   & \textbf{sym int8}   & \textbf{0.0132$\pm$0.0007} & $\pm$\textbf{0.0009} & \textbf{63.2$\times$} \\
8   & asym uint8 & 0.0130$\pm$0.0007 & $\pm$0.0009 & 63.6$\times$ \\
16  & sym int8   & 0.0137$\pm$0.0010 & $\pm$0.0013 & 60.6$\times$ \\
32  & sym int8   & 0.0157$\pm$0.0008 & $\pm$0.0010 & 59.7$\times$ \\
64  & sym int8   & 0.0171$\pm$0.0006 & $\pm$0.0008 & 58.5$\times$ \\
128 & sym int8   & 0.0178$\pm$0.0006 & $\pm$0.0007 & 55.5$\times$ \\
256 & sym int8   & 0.0160$\pm$0.0005 & $\pm$0.0006 & 56.2$\times$ \\
\bottomrule
\end{tabular}
\end{table}

\begin{table}[t]
\centering
\caption{Infer-only Latency, Batch=1 (Numba LUT). Baseline: 158.9$\pm$28.7 ms/sample.}
\label{tab:latency_batch1}
\scriptsize
\begin{tabular}{rlccc}
\toprule
$L$ & Quant. & ms/sample ($\mu\pm\sigma$) & CI$_{95\%}$ & Speedup \\
\midrule
2   & sym int8   & 0.0250$\pm$0.0051 & $\pm$0.0063 & 5759$\times$ \\
4   & sym int8   & 0.0250$\pm$0.0009 & $\pm$0.0011 & 5885$\times$ \\
\textbf{8}   & \textbf{sym int8}   & \textbf{0.0250$\pm$0.0008} & $\pm$\textbf{0.0010} & \textbf{6333$\times$} \\
16  & sym int8   & 0.0260$\pm$0.0008 & $\pm$0.0010 & 6042$\times$ \\
32  & sym int8   & 0.0260$\pm$0.0008 & $\pm$0.0010 & 6070$\times$ \\
64  & sym int8   & 0.0310$\pm$0.0051 & $\pm$0.0063 & 5066$\times$ \\
128 & sym int8   & 0.0320$\pm$0.0050 & $\pm$0.0062 & 4963$\times$ \\
256 & sym int8   & 0.0310$\pm$0.0044 & $\pm$0.0054 & 5107$\times$ \\
\bottomrule
\end{tabular}
\end{table}

\begin{figure*}[t]
    \centering
    \includegraphics[width=\textwidth]{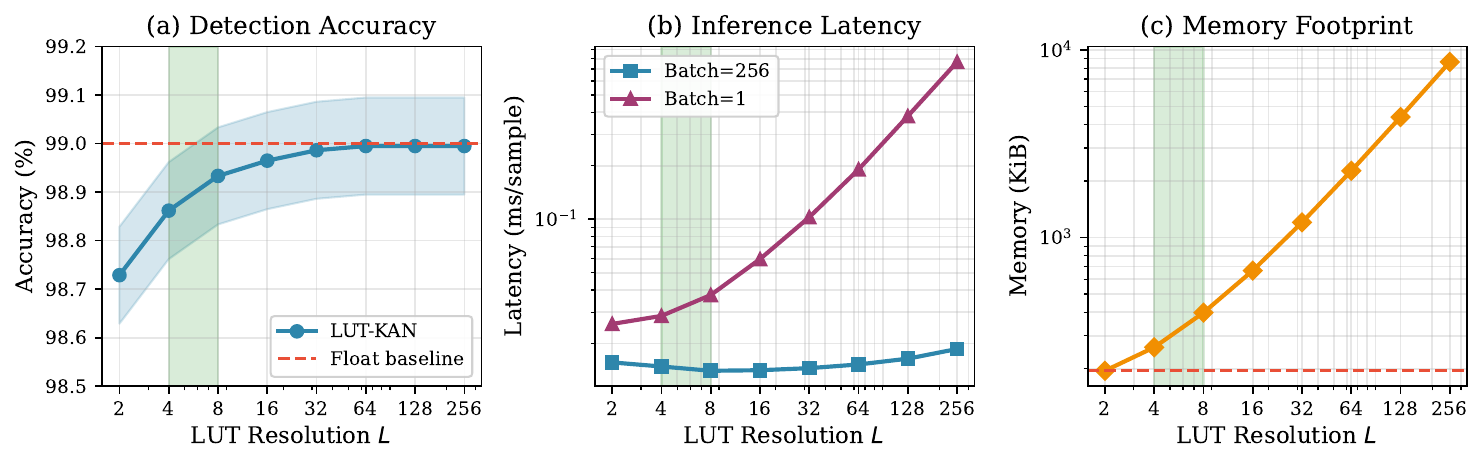}
    \caption{Accuracy-latency-memory trade-off across LUT resolutions. (a) Detection accuracy stabilizes above $L=8$. (b) Inference latency shows optimal speedup at $L=4$--$8$. (c) Memory footprint scales linearly with $L$. Green shaded region indicates Pareto-optimal configurations.}
    \label{fig:tradeoff}
\end{figure*}

Fig.~\ref{fig:tradeoff} presents the comprehensive accuracy-latency-memory trade-off.

\subsection{Memory Footprint Analysis}

Table~\ref{tab:memory} details memory consumption across resolutions.

\begin{table}[h]
\centering
\caption{Memory Footprint by LUT Resolution. Float model: 195.7 KB.}
\label{tab:memory}
\begin{tabular}{ccc}
\toprule
$L$ & \textbf{LUT Size (KiB)} & \textbf{Ratio to Float} \\
\midrule
2 & 195.0 & 1.00$\times$ \\
4 & 260.0 & 1.33$\times$ \\
\textbf{8} & \textbf{390.0} & \textbf{1.99$\times$} \\
16 & 649.8 & 3.32$\times$ \\
32 & 1,169.6 & 5.98$\times$ \\
64 & 2,209.1 & 11.29$\times$ \\
128 & 4,288.1 & 21.91$\times$ \\
256 & 8,446.1 & 43.16$\times$ \\
\bottomrule
\end{tabular}
\end{table}

\begin{figure}[t]
    \centering
    \includegraphics[width=0.9\columnwidth]{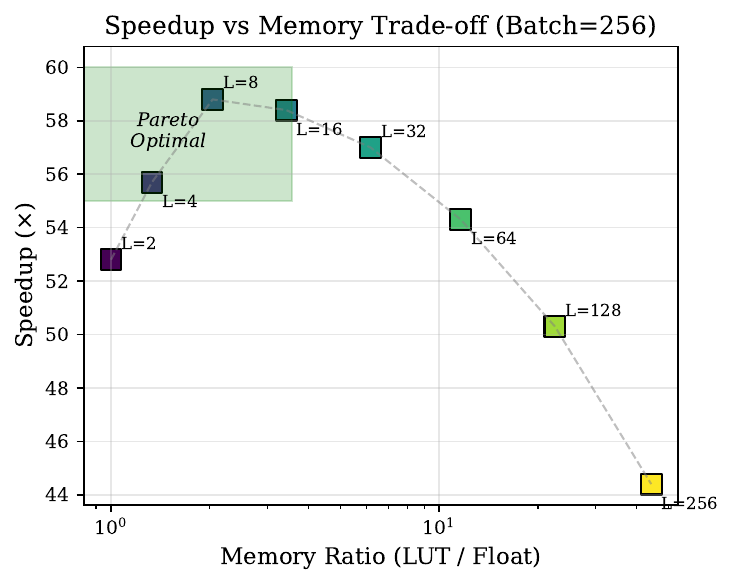}
    \caption{Speedup vs. memory ratio Pareto frontier (Batch=256). Configurations with $L \in \{4, 8\}$ achieve maximum speedup with minimal memory overhead.}
    \label{fig:pareto}
\end{figure}

Fig.~\ref{fig:pareto} shows the Pareto frontier for speedup versus memory ratio.

\subsection{Quantization Scheme Comparison}

\begin{figure}[t]
    \centering
    \includegraphics[width=\columnwidth]{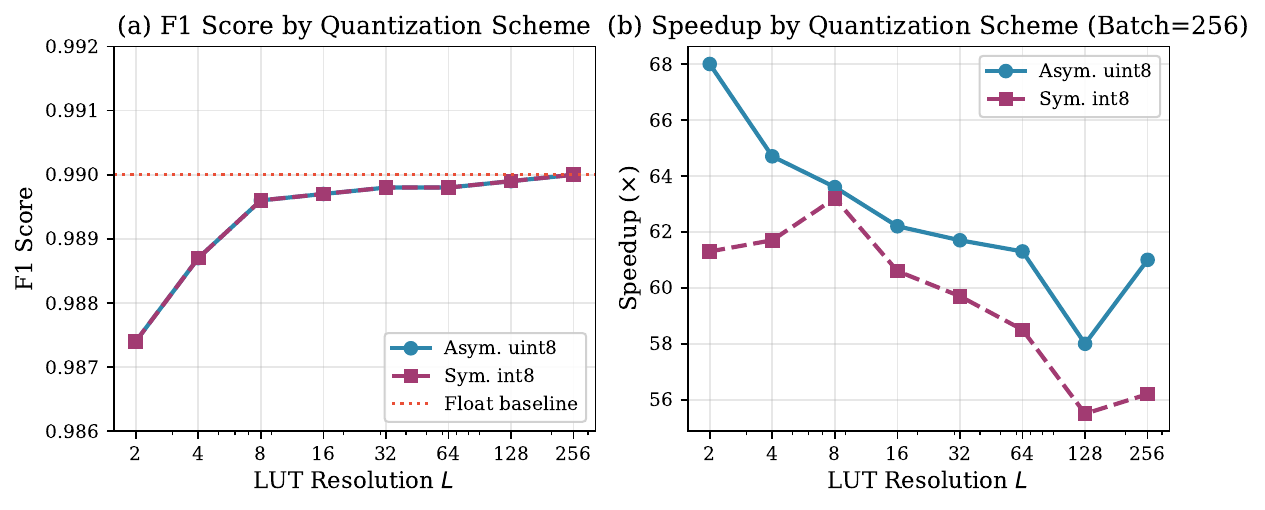}
    \caption{Quantization scheme comparison. (a) F1 scores are indistinguishable. (b) Speedup shows minor variations.}
    \label{fig:quantization}
\end{figure}

Fig.~\ref{fig:quantization} compares symmetric int8 and asymmetric uint8 quantization. Both achieve nearly identical F1 scores, indicating that for DoS detection, the dominant factor is LUT resolution $L$ rather than quantization scheme.

\subsection{OOB Handling and Robustness Analysis}

Real-world deployment may encounter inputs outside the training distribution. We analyze robustness by separately evaluating in-range and out-of-bounds (OOB) samples.

\textbf{OOB Characterization.} In the test set, approximately 2.3\% of feature values fall outside the trained spline domain boundaries after standardization. These OOB values arise primarily from extreme flow statistics (e.g., unusually large packet counts or durations).

Table~\ref{tab:oob} presents results for different boundary handling configurations at $L=4$. The \texttt{zero\_spline} policy zeros the spline contribution for OOB inputs while preserving the base branch, providing graceful degradation. The \texttt{clip\_x} policy clamps inputs to domain boundaries, which may introduce artifacts for extreme values.

\begin{table}[h]
\centering
\caption{Boundary and OOB Policy Ablation at $L=4$ (sym int8). All configurations achieve identical F1 on in-range samples.}
\label{tab:oob}
\scriptsize
\begin{tabular}{llcccc}
\toprule
Boundary & OOB Policy & F1 (all / in-range) & ms/sample (bs=256) & ms/sample (bs=1) \\
\midrule
closed & clip\_x & 0.9887 / 0.9889 & 0.015$\pm$0.006 & 0.029$\pm$0.009 \\
closed & zero\_spline & 0.9887 / 0.9889 & 0.013$\pm$0.001 & 0.025$\pm$0.001 \\
half\_open & clip\_x & 0.9887 / 0.9889 & 0.013$\pm$0.000 & 0.027$\pm$0.005 \\
\textbf{half\_open} & \textbf{zero\_spline} & \textbf{0.9887} / \textbf{0.9889} & \textbf{0.013$\pm$0.001} & \textbf{0.025$\pm$0.001} \\
\bottomrule
\end{tabular}
\end{table}

The minimal F1 difference between configurations indicates that OOB handling is not critical for CICIDS2017 due to the low OOB rate. However, for deployment scenarios with potential distribution drift, we recommend \texttt{half\_open + zero\_spline} as the default robust configuration.

\subsection{Comparative Analysis}

Table~\ref{tab:comparison} positions our approach against recent IoT IDS methods. Direct latency comparison requires caution due to different hardware platforms and measurement protocols.

\begin{table}[h]
\centering
\caption{Comparison with State-of-the-Art IoT IDS. Latency values from original papers; direct comparison requires caution due to different platforms.}
\label{tab:comparison}
\scriptsize
\begin{tabular}{lcccc}
\toprule
\textbf{Method} & \textbf{Acc.} & \textbf{Size} & \textbf{Latency} & \textbf{Platform} \\
\midrule
SecEdge~\cite{awan2025secedge} & 98.7\% & 1.1--1.7 GB & -- & GPU \\
ALNS-CNN~\cite{cherfi2024alns} & 99.85\% & -- & $\sim$15 ms$^{\dagger}$ & CPU \\
NIDS-DA~\cite{kumar2025nidsda} & 99.97\% & 5K params & -- & -- \\
EDA-GWO-XGB~\cite{bikila2025edagwo} & 99.87\% & -- & -- & CPU \\
\midrule
\textbf{LUT-KAN (Ours)} & \textbf{98.96\%} & \textbf{0.4 MB} & \textbf{0.013 ms}$^{\ddagger}$ & \textbf{CPU} \\
\bottomrule
\multicolumn{5}{l}{\scriptsize $^{\dagger}$Per-batch latency, batch size unspecified.} \\
\multicolumn{5}{l}{\scriptsize $^{\ddagger}$Per-sample at batch=256, AMD Ryzen 7 7840HS, single-threaded.} \\
\end{tabular}
\end{table}

Our LUT-KAN achieves competitive accuracy with the smallest memory footprint and lowest reported latency among CPU-based methods. The key advantage is deterministic, sub-millisecond inference suitable for real-time IoT deployment.

\section{Discussion}
\label{sec:discussion}

\subsection{From Baseline KAN to LUT-Compiled Inference}

In our baseline (non-LUT) study, we implemented a lightweight KAN-based IDS for DoS detection with a compact model footprint (50K parameters) and strong predictive performance (99\% accuracy). The key argument was that KAN can achieve high accuracy with a small parameter budget, attractive for IoT/edge environments.

However, the baseline KAN implementation relies on runtime B-spline evaluation and a deep-learning stack (PyTorch/PyKAN). This makes CPU inference sensitive to Python overhead, operator scheduling, and library dependencies, especially in low-batch regimes. The LUT-KAN compilation targets exactly this bottleneck: it replaces spline evaluation by a packed, table-driven execution path (NumPy/Numba), while preserving the trained function shape up to controlled approximation error.

\subsection{Accuracy-Latency-Memory Trade-offs}

Our results establish clear guidelines for LUT resolution selection:

\textbf{$L=2$--$4$}: Suitable for extremely constrained devices. F1 remains above 0.987---sufficient for first-stage filtering in hierarchical detection systems. LUT size approximately equals float model size.

\textbf{$L=8$}: \textbf{Recommended default.} Achieves 98.96\% accuracy with only 2$\times$ memory overhead and maximum speedup (58--63$\times$ at batch=256, 6000$\times$ at batch=1). The sweet spot where further resolution increases yield diminishing returns.

\textbf{$L \geq 16$}: Unnecessary for CICIDS2017; accuracy plateaus while memory grows linearly.

\begin{figure}[t]
    \centering
    \includegraphics[width=0.75\columnwidth]{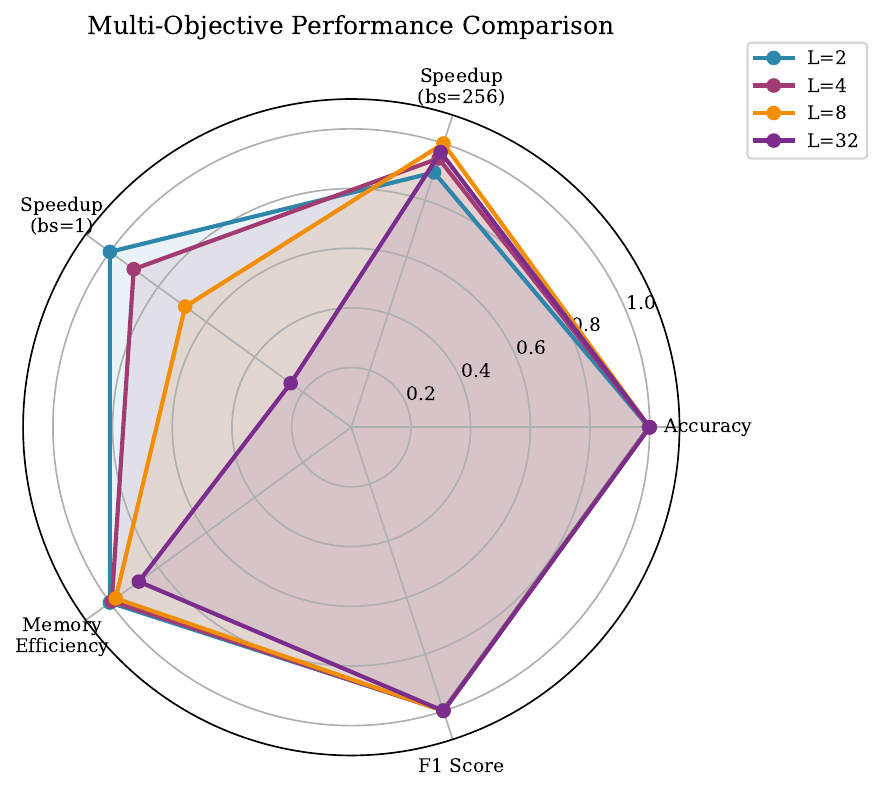}
    \caption{Multi-objective performance comparison across LUT resolutions. $L=8$ achieves the best balance across all metrics.}
    \label{fig:radar}
\end{figure}

Fig.~\ref{fig:radar} provides a multi-objective view comparing selected configurations.

\subsection{Interpreting Recall Changes After LUT Compilation}

It is possible that recall slightly increases after LUT approximation even if accuracy remains unchanged. LUT compilation introduces a small, structured perturbation in the decision function, which can move some borderline samples across the fixed threshold. On imbalanced data, a small shift can increase recall while slightly decreasing precision. We interpret such effects as threshold sensitivity rather than systematic improvement; hence we report ROC-AUC and PR-AUC in addition to thresholded metrics.

\subsection{Deployment Implications}

The LUT compilation approach offers several practical advantages:

\textbf{Deterministic Inference.} Unlike float B-spline evaluation with variable recursion depth, LUT lookup performs a fixed number of operations regardless of input values. This enables precise timing analysis critical for real-time systems.

\textbf{Reduced Dependencies.} Compiled LUT models require only basic array indexing and linear interpolation---eliminating dependencies on numerical libraries for spline evaluation.

\textbf{Hardware Acceleration Potential.} The simple operations map efficiently to SIMD instructions and could be further accelerated with custom hardware or FPGA implementation.

\subsection{Limitations and Future Work}

\textbf{Platform Scope.} Our evaluation uses x86 hardware; ARM-based IoT devices may exhibit different characteristics due to cache hierarchies and SIMD capabilities.

\textbf{Energy Consumption.} We report latency but not direct power measurements. However, the dramatic latency reduction (63$\times$ fewer CPU cycles at batch=256) suggests proportional energy savings, as CPU power consumption correlates strongly with active instruction count on modern processors. For battery-powered IoT devices, this latency-to-energy proxy indicates that LUT-KAN could extend operational lifetime significantly compared to B-spline evaluation. Rigorous energy profiling with hardware power meters on target ARM platforms remains important future work.

\textbf{Single Dataset.} Validation on NSL-KDD, UNSW-NB15, CICDDoS2019 would strengthen generalizability claims.

\textbf{Multi-class Detection.} Our binary formulation could be extended to distinguish attack types, though this may require larger KAN architectures.

Future work will address ARM deployment with direct energy profiling, explore quantization-aware training for improved low-resolution performance, and investigate multi-class KAN architectures for fine-grained attack classification.

\section{Conclusion}
\label{sec:conclusion}

We presented LUT-compiled Kolmogorov-Arnold Networks for lightweight DoS detection on IoT edge devices. Our approach addresses the B-spline evaluation bottleneck by replacing runtime spline computation with precomputed quantized lookup tables and linear interpolation.

On CICIDS2017, our lightweight KAN model (50K parameters, 0.19~MB) achieves 99.0\% detection accuracy. After LUT compilation with resolution $L=8$, the model maintains 98.96\% accuracy while achieving 63$\times$ speedup at batch size 256 and over 6000$\times$ at batch size 1, with only 2$\times$ memory overhead.

The results show a clear accuracy-latency-memory trade-off controlled primarily by $L$: small values provide near-float quality with modest memory growth, while larger values approach lossless approximation at higher memory cost. Symmetric int8 and asymmetric uint8 quantization perform similarly, suggesting that $L$ and OOB handling are the main practical deployment knobs.

LUT-KAN makes KAN-based IDS viable for IoT/edge settings by reducing inference latency and simplifying the execution path. The deterministic latency and minimal resource footprint make this approach particularly suitable for real-time threat detection at the network edge.

\section*{Data Availability}

The source code, trained models, and experimental data are publicly available at: \url{https://github.com/KuznetsovKarazin/kan-dos-detection/tree/lut-v2}

\section*{Acknowledgment}
The author thanks the anonymous reviewers for their constructive feedback.

\bibliographystyle{IEEEtran}
\bibliography{references_ieee}

\begin{IEEEbiography}[{\includegraphics[width=1in,height=1.25in,clip,keepaspectratio]{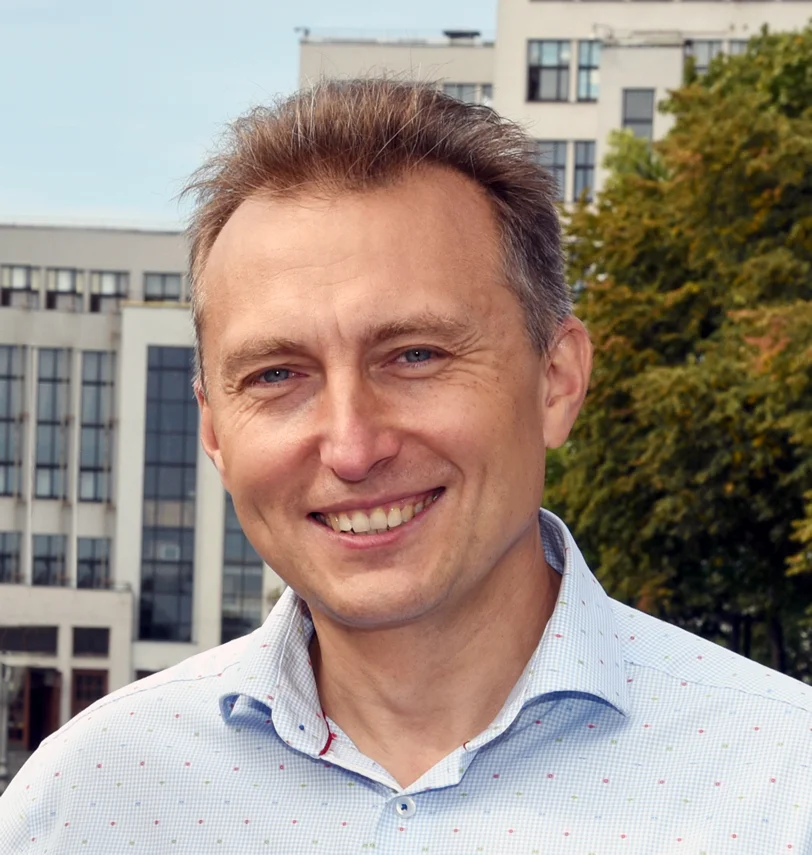}}]{Oleksandr Kuznetsov}
holds a Doctor of Sciences degree in Engineering and is a Full Professor. He is an Academician at the Academy of Applied Radioelectronics Sciences and the recipient of the Boris Paton National Prize of Ukraine in 2021. Additionally, he serves as a Professor at the Department of Theoretical and Applied Sciences at the eCampus University in Italy. He is also a Professor at the Department of Intelligent Software Systems and Technologies at the V. N. Karazin Kharkiv National University, Ukraine. His research primarily focuses on applied cryptology and coding theory, blockchain technologies, the Internet of Things (IoT), and the application of AI in cybersecurity.
\end{IEEEbiography}

\end{document}